\documentclass[lettersize,journal]{IEEEtran}
\usepackage{booktabs}
\usepackage{amsmath,amsfonts}
\usepackage{algorithmic}
\usepackage{algorithm}
\usepackage{array}
\usepackage[caption=false,font=normalsize,labelfont=sf,textfont=sf]{subfig}
\usepackage{textcomp}
\usepackage{stfloats}
\usepackage{url}
\usepackage{verbatim}
\usepackage{graphicx}
\usepackage{cite}
\begin{document}

\title{Dense Optimizer : An Information Entropy-Guided \\Structural Search Method for Dense-like Neural Network Design}

\author{Tianyuan Liu, Libin Hou, Linyuan Wang, Xiyu Song, Bin Yan\\
Information Engineering University, ZhengZhou,China
\thanks{This paper was produced by the IEEE Publication Technology Group. They are in Piscataway, NJ.}
\thanks{Manuscript received April 19, 2021; revised August 16, 2021.}}

\markboth{Journal of \LaTeX\ Class Files,~Vol.~14, No.~8, August~2025}%
{Shell \MakeLowercase{\textit{et al.}}: A Sample Article Using IEEEtran.cls for IEEE Journals}


\maketitle

\begin{abstract}
Dense Convolutional Network has been continuously refined to adopt a highly efficient and compact architecture, owing to its lightweight and efficient structure. However, the current Dense-like architectures are mainly designed manually, it becomes increasingly difficult to adjust the channels and reuse level based on past experience. As such, we propose an architecture search method called Dense Optimizer that can search high-performance dense-like network automatically. In Dense Optimizer, we view the dense network as a hierarchical information system, maximize the network’s information entropy while constraining the distribution of the entropy across each stage via a power-law, thereby constructing an optimization problem. We also propose a branch-and-bound optimization algorithm, tightly integrates power-law principle with search space scaling to solve the optimization problem efficiently. The superiority of Dense Optimizer has been validated on different computer vision benchmark datasets. Specifically, Dense Optimizer completes high-quality search but only costs 4 hours with one CPU. Our searched model DenseNet-OPT  achieved a top-1 accuracy of 84.3\% on CIFAR-100, which is 5.97\% higher than the original one. 
\end{abstract}


\section{Introduction}
\IEEEPARstart{I}{n} the field of computer vision, searching for superior and lightweight network architecture will never be an outdated task. Series of work like AlexNet ~\cite{krizhevsky2012imagenet}, VGG~\cite{simonyan2014very}, ResNet~\cite{he2016deep}, and DenseNet~\cite{huang2019convolutional} has improved the effectiveness of neural networks and our understanding of network design.
Since the proposal of DenseNet, its design concept has been widely applied to the design of various advanced backbone network models. Networks based on improvements to Dense, such as Channel Cross-Linked Dense Convolutional Networks~\cite{chen2017dual}, lightweight DenseNet structure, dense unit, dense connection mode, and attention mechanism, have achieved significant accomplishments in tasks such as EEG emotion detection, medical analysis, etc.~\cite{zhou2022dense,yang2018convolutional,lodhi2019multipath,chen2021multipath}.

However, with the expansion of model scale, it becomes increasingly difficult to adjust the network connection channel design and reuse level based on past experience. So far, Neural Architecture Search(NAS) have provided convenience for structural design by constructing high-performance networks using reinforcement learning or gradient-based algorithms within a given fixed search space~\cite{guo2020single}~\cite{liu2018darts}. But to some extent, automatic search does not equal a good automatic design method. These previous NAS works cannot skillfully control aspects such as the convolution kernel and the number of channels. Meanwhile NAS methods require the training and evaluation of a large number of networks during the search phase, which is both time-consuming and computationally intensive.

The bilevel optimization of privious NAS, like differentiable architecture search(DARTS)\cite{liu2018darts}, brings serious computational overhead problem. Our solution is to separate the structural parameter search and weight parameter training of Dense, decouple the bilevel optimization into a two-step optimization, and transform the design of structural hyperparameters (such as the size of the convolution kernel, the number of channels, etc.) into an optimization problem. Information entropy is an effective tool to describe the network representation ability\cite{Jaynes1957InformationTA}, which can be used as the main optimization objective to search Dense structures. After the Dense structural parameters are optimized, the weight parameters will be trained to obtain the final high-performance model.

In practical terms, we propose Dense Optimizer, a universal structural parameters optimizer for dense-like network structures. 
Dense Optimizer establishes an optimization model to maximize the structural information entropy of the dense backbone network by searching for the optimal configuration of network depth and width and kernel size. And it incorporate a power-law distribution as an evaluation metric, and naturally embedding it into the optimization model. Besides, we propose a branch-and-bound optimization algorithm based on search space scaling to solve the problem. Utilizing traditional dense-BC modules, the models designed by Dense Optimizer are comparable with CNN models of the same size and FLOPs. The following are the key contributions of this work:
\begin{itemize}
  \item Dense Optimizer is proposed as a dense architecture search method. The search process is transformed into an optimization problem to construct Dense model efficiently.
  \item Maximizing the network’s information entropy under multi-scale entropy power-law distribution principle is proposed to conduct the optimization model.
  \item A branch-and-bound optimization algorithm is proposed, tightly integrate power-law principle with search space scaling.

\newpage
  
  \item Dense Optimizer is specifically designed for dense-like architecture, and achieve significant improvements across different datasets.
\end{itemize}

\section{Related Work}
\subsection{Design of DenseNet}
DenseNet has garnered widespread attention and research interest in the field of computer vision~\cite{huang2017densely}. Owing to its excellent performance and flexibility, a multitude of manual designs in recent years have leveraged to refine and augment the DenseNet architecture \cite{li2021transfer}\cite{srivastva2021plexnet}. However, these designs still heavily rely on on human expertise and lack principles for structural design guidance~\cite{zhou2022dense}. When selecting hyperparameters such as channel growth rate and convolution kernels, extensive experimentation and tuning are often required\cite{zhao2019deep}. Dense Optimizer provides an efficient and automatic structural search method for Dense-like networks and promote its performance at the same time. 

\subsection{Neural Architecture Search}
Neural Architecture Search(NAS) have been greatly studied during the past few years to automatically design more effective architectures. Popular NAS algorithm options include genetic algorithms \cite{liu2021survey,ying2020neural,real2017large,xie2017genetic}, reinforcement learning (RL) \cite{tan2019mnasnet,zoph2016neural}, differentiable search \cite{cai2018proxylessnas,liu2018darts} and many other types of optimization algorithms, e.g.,  Bayesian optimization and particle swarm optimization \cite{eriksson2021latency,nistor2022intelliswas,white2021bananas}. As a classic differentiable architecture search method, DARTS is the most widely attention-grabbing algorithms. It models the architecture design as a bi-level optimization problem, which requires training vast numbers of candidate networks to inform the search process, and often leads to high computational cost. 

Recent NAS work Mellor et al.~\cite{mellor2021neural} start to explore indicators that can predict a network’s performance without training. Moreover, Xuan et al.~\cite{shen2023deepmad} endeavor to decouple network weights from network architecture, focusing on the discovery of improved design principles through the exploration of structural parameters alone. These methods have represented an effective exploration in neural network architecture search that decouples structural parameters and weights, but still lack good design guidelines. Dense Optimizer conducts further in-depth research. Dense Optimizer circumventing the time-consuming issue of bi-level optimization, and uses information entropy as a design criterion to obtain a better structure by solving an optimization problems.

\subsection{Mathematical Architecture Design}

Information theory is a powerful tool for studying complex systems such as deep neural networks. Recently, mathematical architecture design(MAD) is proposed~\cite{shen2023deepmad}. 
Unlike the existing hyperparameter optimization methods~\cite{bischl2023hyperparameter,yu2020hyper,yu2020learning}, MAD does not require any model training during optimization, allowing for the acquisition of optimized network structures within minutes. It maximizes the network entropy with three empirical guidelines and demonstrate an advancement in designing network structures using mathematical programming. However, MAD unable to characterize the information flow structure of the concatenation operation and accurately estimate the information entropy of the dense network structure. And its three empirical guidelines do not have strong theoretical foundation. Dense Optimizer not only specifically addresses these issues, but also constraints the distribution of entropy at different scales with power-law and proposes a new optimization model for dense-like network architectures.

\section{Dense based architecture Optimizer}
\label{sec:MAD}

In this section, we introduce the core architecture of Dense Optimizer. Specifically, we consider the deep neural networks as a continuous information processing system. We provide a definition of structural entropy's effectiveness and extend it from Multi-Layer Perceptron(MLP) to networks with dense connections. Then, we propose an optimization model to study the architectural design of DenseNet. To formulate the optimization problem for DenseNet, we first define the entropy that governs its expressiveness, followed by the power-law constraints that regulate the efficacy of the multi-scale entropy distribution. Ultimately, We provided a precise optimization model and proposed the corresponding branch-and-bound optimization algorithm.

\subsection{Entropy of DenseBlock}

The principle of maximum entropy is one of the most widely applied principles in information theory~\cite{chan2022redunet}. Some previous works have also attempted to establish a connection between entropy and the structure of neural networks. Here, we provide a re-derivation, and give the entropy upper bound of DenseBlock. Suppose that in an L-layer MLP f(·), the i-th layer has $\mathrm{w}_i$ input channels and $\mathrm{w}_{\mathrm{i}+1}$ output channels. The output $\mathrm{x}_{\mathrm{i}+1}$ and the input $\mathrm{x}_i$ are connected by $\mathrm{x}_{\mathrm{i}+1}=\mathrm{M}_{\mathrm{i}} \mathrm{x}_i$ , where $M_i \in R^{w_{i+1} \times w_i}$ is trainable weights. Then the structural parameters define how the input $\mathrm{x}_i$ propagates inside the network, which is capable of being depicted.

For a DenseBlock with L layers, we consider the information reuse caused by the network's dense connections and the impact of information distribution brought about by the concatenation operation. Then we obtain the information entropy of the basic block of the dense network, defined in Proposition 1:

{\bf Proposition 1.}The normalized Gaussian entropy upper bound of the DenseBlock f(·) is
\begin{equation}
H_f=\mathrm{w}_L \log \left(\mathrm{w}_0^{\mathrm{L}} * \mathrm{i} !\right),
\label{eq:1}
\end{equation}
where the $\mathrm{w}_L$ is the width of $L$-th layer, and $\mathrm{w}_0$ is the initial width of the DenseBlock.
The whole derivation is given in Appendix A. The entropy measures the expressiveness of a DenseBlock. Following the Principle of Maximum Entropy \cite{csiszar2004information,kullback1997information}, we propose to maximize the entropy of DenseBlock under given computational budgets. When calculating the precise information entropy of a dense block, the number of the $i$-th dense layer, the input channels is $\mathrm{c}_i$, the number of output channels is $\mathrm{c}_{\mathrm{i}+1}$, and the kernel size is $\mathrm{k}_i$. Consequently, the "width" of a dense block layer is projected as $\mathrm{c}_i\mathrm{k}_i^2$ in (\ref{eq:1}). Therefore, for a feature map with a resolution of $\mathrm{r}_i \times \mathrm{r}_i $, the entropy of a DenseBlock with L layers is defined by
\begin{equation}
H_f=\log \left(\mathrm{r}_L^2\mathrm{c}_{\mathrm{L}+1}\right) \log \left(\left(\mathrm{c}_i\mathrm{k}_i^2\right)^{\mathrm{L}} * \mathrm{i} !\right).
\label{eq:2}
\end{equation}

Inspired by \cite{hyvarinen2009natural}, taking logarithms can better formulate the ground-truth entropy for natural images.

\subsection{Effectiveness Defined in DenseBlock}

Inspired by previous research, an infinitely deep network will become hard to train unless it meets particular structural requirements. Therefore, in Dense Optimizer, we propose to control the depth of the dense-like network to facilitate gradient flow throughout the entire architecture effectively. Typically, the depth and width of a network are relative; thus, the effectiveness of a network with $\mathrm{L}$ layers, where each layer possesses the same width $\mathrm{W}$, can be defined as follows:
\begin{equation}
\rho=\mathrm{L} / \mathrm{W}.
\end{equation}

Normally, the width $\mathrm{w}_i$  of each layer can be different. So the average width of an $\mathrm{L}$ layer network f(·) is defined by

\begin{equation}
\overline{\mathrm{w}}=\left(\prod_{i=1}^L w_i\right)^{1 / L}=\exp \left(\frac{1}{L} \sum_{i=1}^L \log w_i\right).
\end{equation}

But in DenseBlock, each layer connects to all the previous layers. This results in a relatively steady increase in the number of parameters with each layer. The growth rate $\mathrm{K}$ is always the same, and much smaller than the number of input channels. So the average of the width can be defined as:
\begin{equation}
\overline{\mathrm{w}}=\mathrm{w}_0+K/2 \approx \mathrm{w}_0.
\end{equation}

So a DenseBlock has L-layers with $\mathrm{w}_0$ input width and growth rate $\mathrm{K}$, the effectiveness is defined by 
\begin{equation}
\rho=\mathrm{L} / \mathrm{w}_0.
\end{equation}

\subsection{Power law in entropy distribution}
In information theory, entropy reflects the amount of information or uncertainty within a system. Higher information entropy implies the system is more dispersed and diverse~\cite{saxe2019information}. For multi-stage neural networks, they are akin to independent information systems. To eliminate the discrete distribution of information entropy and uncertainty across these systems, it is essential to propose a multi-scale entropy distribution under mathematical constraints.
According to highly optimized tolerance(HOT)\cite{carlson1999highly}, when a complex system is in a HOT state, the system will satisfy power-laws, that is, a global optimization process can lead to power-law distributions: inputs with characteristic scales, after undergoing a global system’s “output” optimization process, can produce outputs with power-law characteristics\cite{Tyagi2016Optimal}.

\begin{figure}[!t]
  \centering
   \includegraphics[width=2.5in]{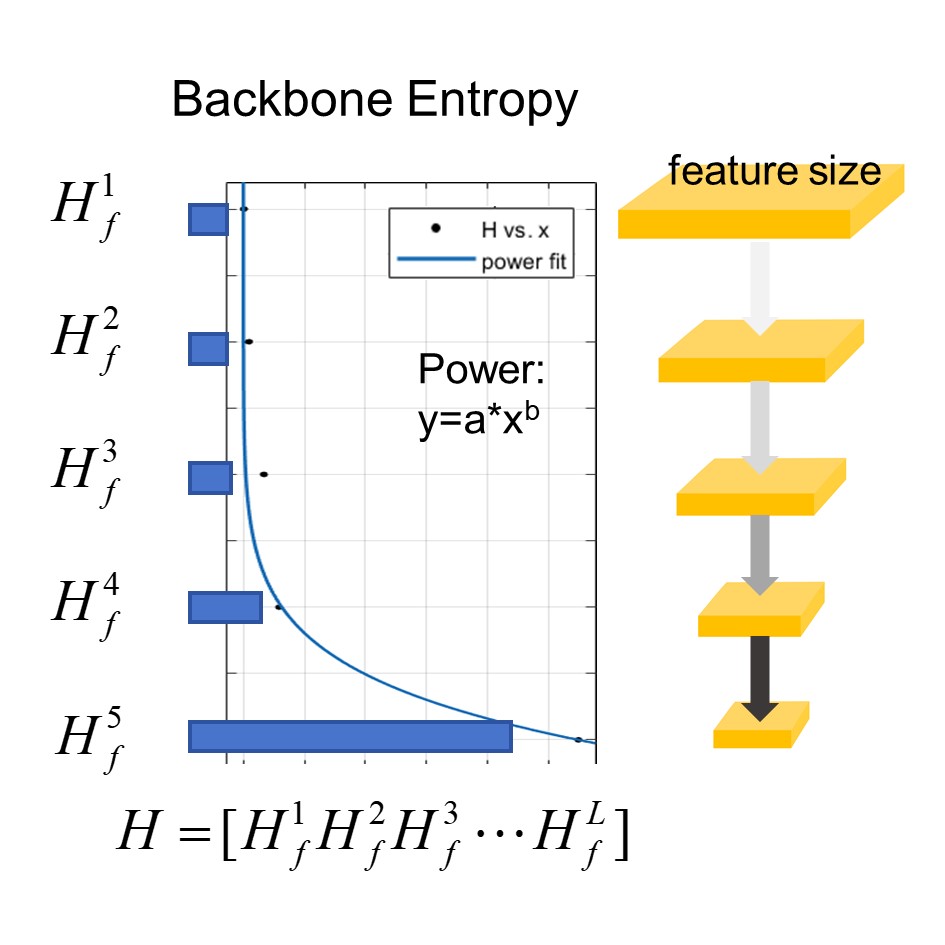}
   \caption{Visualization of the multi-scale entropy power-law distribution, which is based on the statistical results of dense backbone. The distribution of a dense backbone entropy under different feature size consistents with the power-law function}
   \label{fig:entropyl}
\end{figure}

Based on extensive experimental statistics, We find that the information entropy distribution of dense network follows a power-law distribution, see figure \ref{fig:entropyl}.

To reinforce this constraint, We propose a power-law in entropy distribution, and we use a two-parameter fitting function with parameters 'a' and 'b' to optimize the distribution. The object is to maximize the value of 'a' and minimize the value of 'b' under the same fitting parameter settings.

Here, we provide specific definitions. Following (\ref{eq:2}), The cumulative entropy distribution sequence at the current stage is:
\begin{equation}
H=\left[H_f^1 H_f^2 H_f^3 \cdots H_f^L\right].
\end{equation}

The fitting expression of this sequence under the power law function is as follows:
\begin{equation}
H=\mathrm{a}*\mathrm{M}_i^\mathrm{b},
\end{equation}
where a, b are power index parameters, and $\mathrm{M}_i$ represent the i th stages. So the optimization target S is:
\begin{equation}
S=\mathrm{a}-\mathrm{b}.
\end{equation}

Subsequently, we establish optimization constraints to achieve the objective of maximizing a while minimizing b.

\subsection{Optimization model and Solutions}
We gather everything together and present the final optimization model for Dense Optimizer.
Suppose that we aim to design an L-layer dense-like model $F$ with $\mathrm{M}$
stages. The entropy of the $\mathrm{i}$-th stage is denoted as $\mathrm{H}_i$ defined in (\ref{eq:2}). We propose to optimize \{$\mathrm{c}_i$,$\mathrm{k}_i$,$\mathrm{L}_i$\} via the following optimization problem:

\begin{equation}
\begin{array}{ll}
\max _{w_i, L_i} &\sum_{i=1}^M \alpha_{\mathrm{i}} H_i+\beta * S, \\
\text { s.t. } & L_i/\mathrm{W}_i \leq \rho, \\
& \text { FLOPs }[f(\cdot)] \leq \text { budget, } \\
& \text { Params }[f(\cdot)] \leq \text { budget, } \\
& w_1 \leq w_2 \leq \cdots \leq w_L .
\end{array}
\end{equation}
where $\alpha_i$ is the weights of entropies at different scales, $\beta$ is a tuning coefficient. $\rho$ controls the effectiveness of the network whose value is usually tuned in range $[10, 20]$.

Integrating the complexity of above optimization problem and the particularity of the candidate solution space of network structure, we propose a branch-and-bound algorithm to integrate power-law in algorithm \ref{alg:BBO Framework}, to achieve efficient optimization search. By combining the properties of power-law distribution, we decompose the optimization problem into sub-optimization problems, gradually narrowing the search space and finding the optimal solution through searching.

Specifically, we first score the initial densely connected architecture through information entropy representation. Then, during the search process, we relax the initial network structure while employing regional reduction techniques during the relaxation process. Based on the network information entropy at a certain stage, we calculate the entropy space that conforms more to the power law distribution, and prune the search space. In the process of  iterative optimization, we always retain the optimal solution.

\begin{algorithm}[H] 
\caption{ Branch and Bound Method for to-Fine optimization} 
\label{alg:BBO Framework} 
\begin{algorithmic}[0] 
\REQUIRE ~~ \\ 
Search space $S$, inference budget $B$, maximal depth $L$\\ total number of iterations $T$,
evolutionary population size $N$, initial structure $ {F}_0 $, fine-search flag Flag.\\
\ENSURE ~~ 
Dense optimized backbone F*.\\
\STATE Initialize population P =  $ {F}_0 $, Flag=False.
\FOR{$t = 1,2,\dots,T$} 
\STATE Calculate the Network Information Entropy at Each Stage
\STATE Conduct a Mathematical Optimization
\STATE Compute the ideal Information Entropy Under the Power-Law Distribution
\STATE Adjusting the Search Space at Each Stage
\STATE Performing internal Mathematical Optimization for each stage
\STATE Remove networks of the smallest entropy if the size of $P$ exceeds $B$
\ENDFOR
\RETURN Return $F*$ , the network of the highest entropy in P
\end{algorithmic}
\end{algorithm}

\section{Experiments and Results}
\label{sec:Experiments and Results}
In this section, we first describe the detailed settings for search optimization using Dense Optimizer. Then, the optimized dense network is trained, and the training settings are introduced in detail in subsection 4.2, and tested on the CIFAR-10, CIFAR-100, SVHN datasets. Meanwhile, the performance of the optimized structure is compared with the classic ResNet and DenseNet, and ablation experiments are conducted in Section 4.4 to verify the effectiveness of the multi-scale information entropy power-law distribution.

\subsection{Search Settings}
In Dense Optimizer, the number of search populations N is 256, and the number of iterations is 500,000. The classic densenet121 is used as the initial backbone network. Search space parameters: the number of input and output channels for each block, convolution kernel size: [3,5,7], budget layers=130, maximum growth rate: [12,24,40]. Meanwhile the optimization problem is solved on CPU device.

\subsection{Training Settings}
Following previous works , SGD optimizer with momentum 0.9 is adopted to train the dense models. The weight decay is 5e-4 for CIFAR dataset . The initial learning rate is 0.1 with batch size of 32. We use cosine learning rate decay with 5 epochs of warm-up. The number of training epochs is 100 for CIFAR-100. All experiments use the following data augmentations : mix-up, label-smoothing , random erasing, random crop/resize/flip/lighting, and Auto-Augment .

\subsection{Result}
As shown in table \ref{tab:result} and table \ref{tab:result-nas}, after optimization with Dense Optimizer, the performance of the dense network on the CIFAR-100 dataset significantly surpassed the original network. Given diverse latency budgets, our method outperforms the compared NAS methods in terms of the accuracy of the generated/searched architectures. It is solved in CPU device with small search cost.  The optimized model, with a size of 32M, achieved a TOP-1 error rate of 16.96\%, while the larger network of 171M size had a TOP-1 error rate of 15.7\%.

Moreover, we compare the searched architectures on CIFAR-10 and SVHN datasets, showing in table \ref{tab:result-cifar10} and table \ref{tab:result-SVHN}.

The results demonstrate that Dense Optimizer achieves effective results on different datasets. Within the same model size, the accuracy of the optimized models is significantly higher than the original densenet, and they can easily outperform ResNet family.

\begin{table}[!t]
\caption{Results of searched optimized architectures on CIFAR-100\label{tab:result}}
    \centering
    \begin{tabular}{@{}l|c|c|c@{}}
    \toprule
       \textbf{Model} & \textbf{ Parameters} &\textbf{ Max K }& \textbf{ top-1 error}\\
    \midrule
        ResNet~\cite{koonce2021resnet} & 1.7M & - & 27.22 \\
        Wide ResNet~\cite{zagoruyko2016wide} & 36.5M & - & 20.50 \\
        ResNet(pre-activation) \- & 10.2M & - & 22.71 \\ 
        FractalNet~\cite{larsson2016fractalnet} & 38.6M & - & 23.3 \\ 
    \midrule
        DenseNet-BC~\cite{zhou2022dense} & 0.8M & 12 & 24.15 \\ 
        DenseNet & 27.2M & 24 & 23.42 \\ 
        DenseNet-BC(121) & 9.02M & 24 & 19.90 \\ 
    \midrule
        DenseNet-OPT(123) & 24.12M & 24 & 17.74 \\ 
        DenseNet-OPT(129) & 32.60M & 40 & 16.96 \\ 
        DenseNet-OPT(86) & 171.7M & 128 & {\bf 15.70 }\\ 
     \bottomrule
    \end{tabular}
\end{table}

\begin{table}[!t]
\caption{Results of searched optimized architectures on CIFAR-100\label{tab:result-nas}}
    \centering
    \begin{tabular}{@{}p{2cm}|c|c|c@{}}
    \toprule
       \textbf{Method} &{\bf Parameters} &{\bf top-1 error }& {\bf Search Cost}\\
    \midrule
        SNAS~\cite{xie2018snas} & 2.8M & 20.09 & 1.5 GPU-days     \\ 
        DARTS~\cite{liu2018darts} & 3.4M & 21.26 & 0.4 GPU-days \\ 
        ZARTS~\cite{wang2022zarts} & 4.1M & 21.00 & 1.0 GPU-days\\ 
    \midrule
        DenseNet-OPT(123) & 24.12M & 17.74 & 0.2 CPU-days \\  
     \bottomrule
    \end{tabular}
\end{table}

\begin{table}[!t]
  \caption{Results of searched optimized architectures on CIFAR-10\label{tab:result-cifar10}}
    \centering
    \begin{tabular}{@{}l|c|c|c@{}}
    \toprule
       {\bf  Model} &{\bf  Parameters} &{\bf Max K }& {\bf top-1 error}\\
    \midrule
        ResNet & 19.3M & - & 7.93 \\
    \midrule
        DenseNet & 1.0M & 12 & 7.00 \\ 
        DenseNet & 27.2M & 24 & 5.83 \\ 
        DenseNet-BC(250) & 15.3M & 24 & 5.19 \\ 
    \midrule
        DenseNet-OPT(123) & 24.1M & 24 & 3.53 \\ 
     \bottomrule
    \end{tabular}
\end{table}

\begin{table}[!t]
  \caption{Results of searched optimized architectures on SVHN\label{tab:result-SVHN}}
    \centering
    \begin{tabular}{@{}l|c|c|c@{}}
    \toprule
       {\bf  Model} &{\bf  Parameters} &{\bf Max K }& {\bf top-1 error}\\
    \midrule
        ResNet-18 & 11.7M & - & 2.65 \\
    \midrule
        DenseNet-BC & 15.3M & 12 & 1.74 \\ 
    \midrule
        DenseNet-OPT(123) & 24.1M & 24 & 1.49 \\ 
     \bottomrule
    \end{tabular}
\end{table}

\subsection{Ablation Study and Analysis}
In this section, we conducted ablation experiments on the CIFAR-100 dataset with power-law distribution constraints and performed statistical analysis on the information entropy of multi-stage power-law distributions. We utilized traditional dense-BC convolution blocks and networks generated with different information entropy distributions. The network structures found under different max channel growth rates $\mathrm{K}$ [40, 24, 12] were trained, and the fitting values of information entropy distribution at each stage were statistically analyzed. 

\begin{figure*}[!t]
  \centering
\subfloat[]{\includegraphics[width=2.5in]{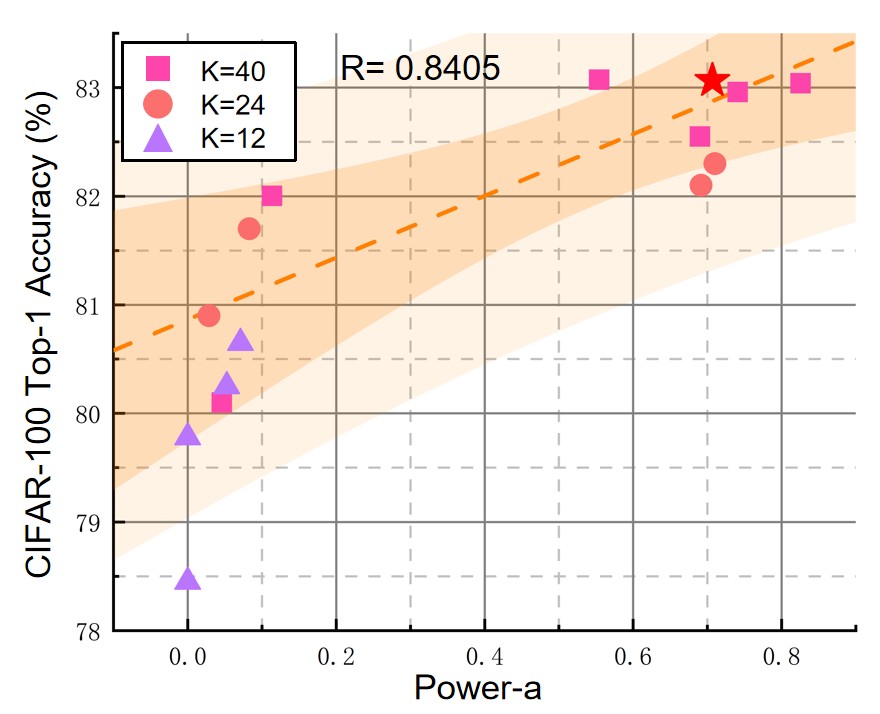}%
\label{fig:power a}}
\subfloat[]{\includegraphics[width=2.5in]{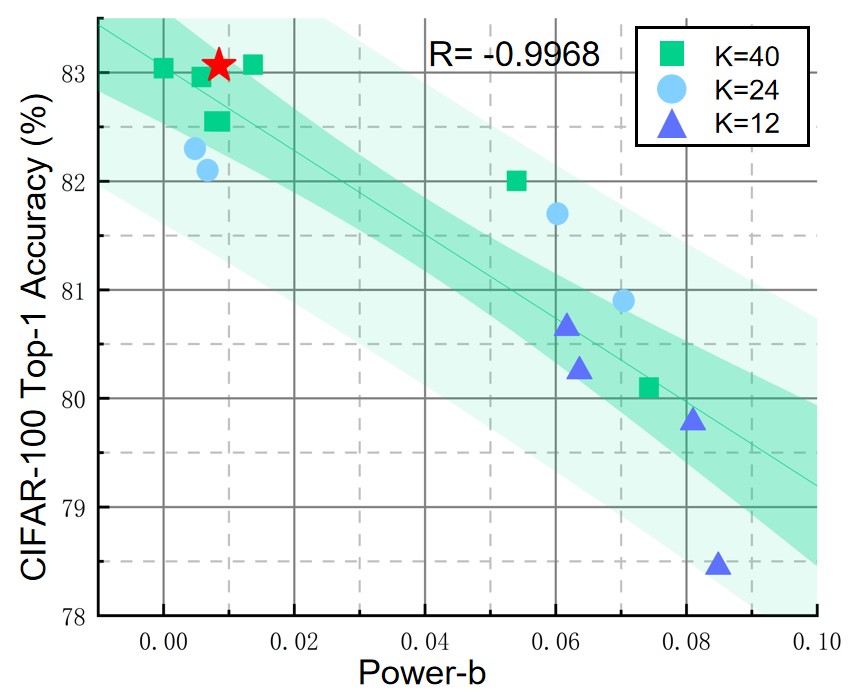}%
\label{fig:power b}}
\hfil
\subfloat[]{\includegraphics[width=2.5in]{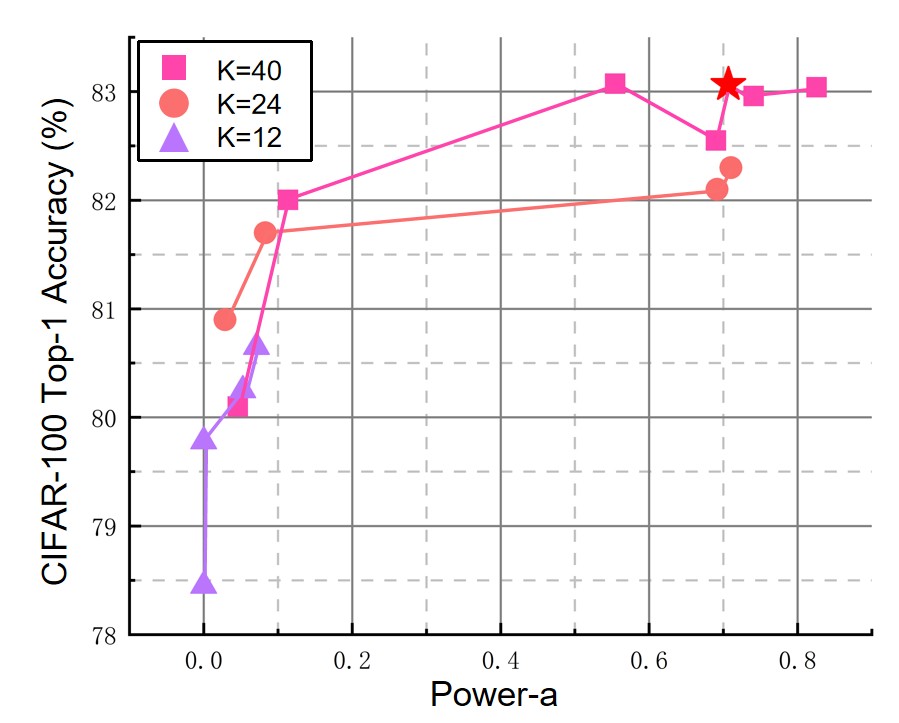}%
\label{fig:power c}}
\subfloat[]{\includegraphics[width=2.5in]{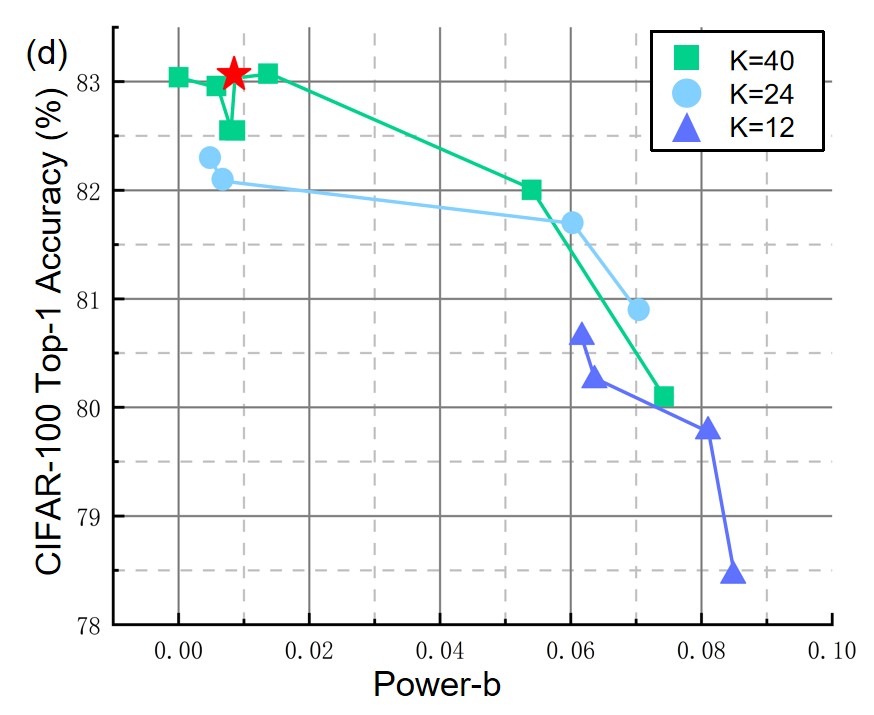}%
\label{fig:power d}}
  \caption{Power fit hyperparameters v.s. top-1 accuracy of each optimized model on CIFAR-100. From left to right: correlation of power fit hyperparameter {\bf a}, {\bf b} against accuracies on CIFAR-100. It reveals strong correlations in both hyperparameters, and maintain consistency across different growth rate settings. }
  \label{fig:power ab}
\end{figure*}

As shown in table \ref{tab:ablation}, controlling the information entropy distribution at each stage can improve the performance of image classification tasks. When the distribution satisfies a power-law distribution, the network can achieve the best model performance in classification tasks. As can be seen in figure \ref{fig:power ab}, the model performance is positively correlated with the power-law distribution hyperparameter a (Pearson correlation coefficient of $0.86$) and negatively correlated with the hyperparameter b (Pearson correlation coefficient of $-0.94$). Therefore, a better performance can be achieved by optimizing the structure of the model through power-law constraints.

\begin{figure*}[!t]
  \centering
  \subfloat[]{\includegraphics[width=2.5in]{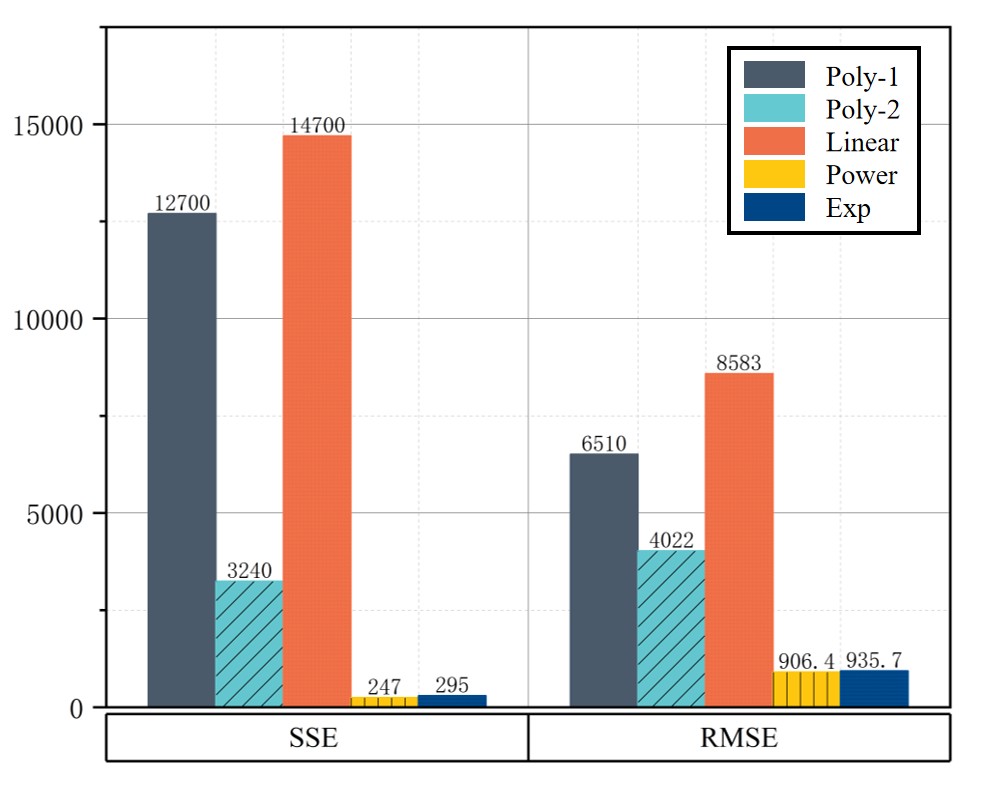}%
\label{fig:power fit a}}
\hfil
\subfloat[]{\includegraphics[width=2.5in]{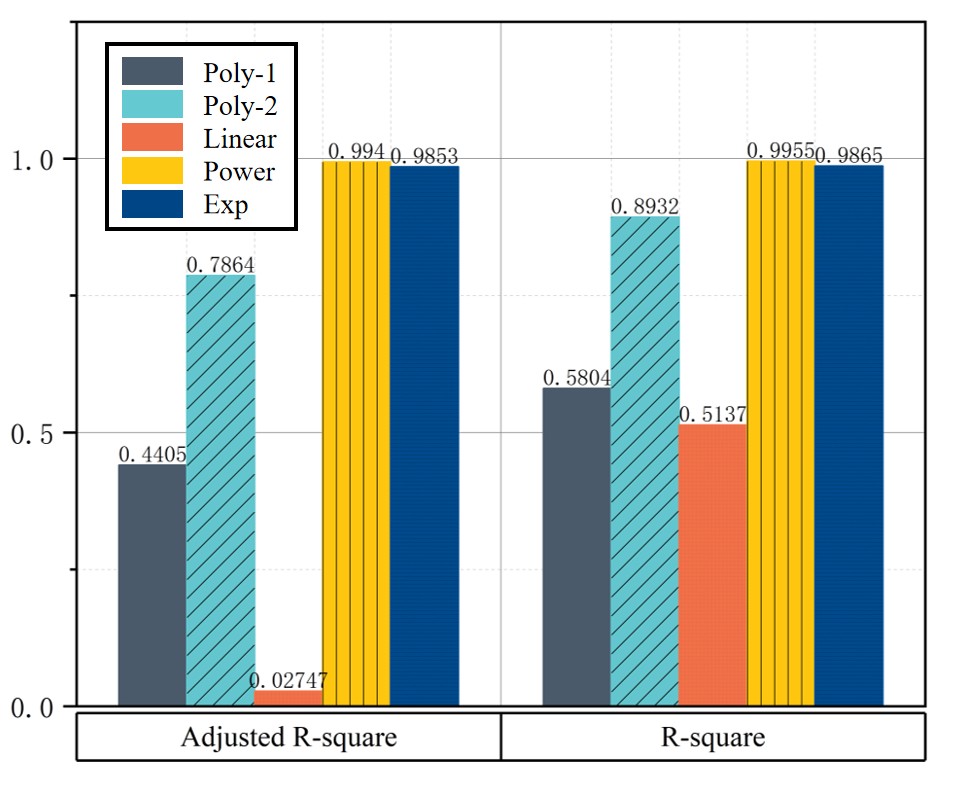}%
\label{fig:power fit b}}
  \caption{From left to right, the figure displays the fitting errors, sum of squared residuals, coefficient of determination, and adjusted coefficient of determination for the high-performance dense network's multiscale information entropy using first-order and second-order polynomials, linear functions, power functions, and exponential functions. Among all the indicators, the power-law function exhibits the best fitting metrics.}
  \label{fig:power fit}
\end{figure*}

We also statistically averaged the information entropy of all optimized dense network stages and fitted them with different functions. And we compared fitting errors, including SSE (Sum of Squared Errors), R-square (Coefficient of Determination), Adjusted R-square, and RMSE (Root Mean Squared Error). Figure \ref{fig:power fit} indicates that, compared to first-order and second-order polynomial functions and exponential functions, the power-law function also has the smallest fitting error and the highest fitting score.

Subsequently, we conducted ablation experiments under power-law constraints. In table~\ref{tab:ablation}, we can observe that the model performance is significantly improved after optimization with the dense Optimizer and the addition of power-law constraint terms. This showcases how the model performance evolves with the strengthening of power-law constraints. The DenseNet, not optimized by power-law, achieved accuracy gains of +0.20\%, +0.57\%, and +2.45\% on SVHN, CIFAR-10, and CIFAR-100 respectively. Meanwhile, the DenseNet optimized under power-law achieved accuracy gains of +0.23\%, +1.66\%, and +2.94\% on SVHN, CIFAR-10, and CIFAR-100 respectively.

\begin{table}[!t]
 \caption{Error rate on the CIFAR and SVHN dataset, * marked are our own test results\label{tab:ablation}}
    \centering
    \begin{tabular}{@{}l|c|c|c@{}}
    \toprule
        {\bf Dataset} & {\bf Original}& {\bf Optimized }& {\bf Power-optimized} \\ 
    \midrule
        SVHN & 1.74 & 1.54* & 1.49* \\ 
        CIFAR-10 & 5.19 & 4.62* & 3.53* \\ 
        CIFAR-100 & 19.90* & 17.45* & 16.96* \\ 
    \bottomrule
    \end{tabular}
\end{table}

Moreover, $\beta$ parameter is crucial in tuning the model. It controls the relationship between the power-law distribution and the magnitude of information entropy. Our results indicate that when $\beta$ is set to 0.1, there is an optimal balance between information entropy and the power-law constraint, leading to the best model performance (See table \ref{tab:ablation-beta}).

\begin{table}[!t]
    \caption{Error rate on the CIFAR dataset with different $\beta$, * marked are our own test results\label{tab:ablation-beta}}
    \centering
    \begin{tabular}{@{}l|c|c|c|c@{}}
    \toprule
       {\bf Dataset} & {\bf $\beta=0$} & {\bf $\beta=0.001$}  & {\bf $\beta=0.1$} & {\bf $\beta=10$} \\
    \midrule
        CIFAR-100 & 19.90* &16.95*  & 16.92* &18.00* \\ 
    \bottomrule
    \end{tabular}
\end{table}

Overall, the analysis proves that there is a strong correlation between hyperparameters {\bf a}, {\bf b} and model performance. The ablation result further validates the significance of power-law constraints in model design. And we analyze the differences between different distribution functions to show the fitness of power. Ultimately, we propose an optimal hyperparameter $\beta$ to achieve the best balance of entropy distribution.

\section{Conclusion}
\label{sec:Conclusion}
In this paper, we propose a dense architecture search method, Dense Optimizer, which can achieve automatic network structure design under mathematical optimization and improve network performance. Dense Optimizer decouples network weights from network architecture and maximizes network entropy while maintaining the distribution of network structure entropy under power-law constraints. We show that Dense Optimizer can design models comparable to modern CNN models using only traditional dense-BC convolutional blocks, proving the powerful capabilities of Dense Optimizer to release the potential of traditional DenseNet models. Furthermore, Dense Optimizer can be applied to the design of other dense-like networks and the power-law distribution characteristic of structural information entropy provides considerable insight for models with multi-scale features.

\section*{Acknowledgments}
This work was supported by the Major Projects of Technological Innovation 2030 of China(Grant number 2022ZD0208500) and the National Natural Science Foundation of China under Grant No.62271504.


\bibliographystyle{IEEEtran}
{
    \small

}

\vfill

\end{document}